\documentclass[sigconf, nonacm, natbib=true]{acmart}
\usepackage[american]{babel}
\usepackage[T1]{fontenc}
\usepackage{natbib}
\usepackage{graphicx}
\usepackage{booktabs}
\usepackage{multirow}
\usepackage{makecell}
\DeclareGraphicsExtensions{.pdf,.ai,.jpg,.png}
\setkeys{Gin}{pagebox=artbox}% Alternative for \pdfpagebox 5 in preceding line.
\graphicspath{{./}}

\newcommand{\Ni}{(1)~}
\newcommand{\Nii}{(2)~}
\newcommand{\Niii}{(3)~}

\begin{document}

\def\bstitle{The Clickbait Challenge 2017:\texorpdfstring{\\}{ }Towards a Regression Model for Clickbait Strength\gdef\bstitle{The Clickbait Challenge 2017: Towards a Regression Model for Clickbait Strength}}

\title{\bstitle}

\author{Martin Potthast}
\affiliation{\institution{Leipzig University}}
\email{martin.potthast@uni-leipzig.de}

\author{Tim Gollub}
\affiliation{\institution{Bauhaus-Universit{\"a}t Weimar}}
\email{tim.gollub@uni-weimar.de}

\author{Matthias Hagen}
\affiliation{\institution{Martin-Luther-Universit\"at Halle-Wittenberg}}
\email{matthias.hagen@informatik.uni-halle.de}

\author{Benno Stein}
\affiliation{\institution{Bauhaus-Universit{\"a}t Weimar}}
\email{benno.stein@uni-weimar.de}

\begin{abstract}
Clickbait has grown to become a nuisance to social media users and social media operators alike. Malicious content publishers misuse social media to manipulate as many users as possible to visit their websites using clickbait messages. Machine learning technology may help to handle this problem, giving rise to automatic clickbait detection. To accelerate progress in this direction, we organized the Clickbait Challenge~2017, a shared task inviting the submission of clickbait detectors for a comparative evaluation. A total of 13~detectors have been submitted, achieving significant improvements over the previous state of the art in terms of detection performance. Also, many of the submitted approaches have been published open source, rendering them reproducible, and a good starting point for newcomers. While the 2017~challenge has passed, we maintain the evaluation system and answer to new registrations in support of the ongoing research on better clickbait detectors.
\end{abstract}

\maketitle

\section{Introduction}

This paper reports on the results of the Clickbait Challenge~2017.%
\footnote{\url{https://clickbait-challenge.org}}
The main goal of the challenge was to kickstart research and development on the novel task of clickbait detection in social media~\cite{stein:2016b}. The term ``clickbait'' refers to social media messages that are foremost designed to entice their readers into clicking an accompanying link to the posters' website, at the expense of informativeness and objectiveness. Typical examples include exaggerated statements, such as ``You won't believe \ldots'', ``\ldots\ will change your life'', or ``\ldots\ will blow your mind'', as well as unnecessary omissions of informative details as in ``{\em This} actor fell \ldots'', ``{\em This} city started \ldots'', ``{\em This} fact about \ldots''.

Spreading content through clickbait has meanwhile become an established practice on social media, even among reputed news publishers~\cite{stein:2016b}. Clickbait works: it has a measurable effect on page impressions, which explains its widespread usage. But clickbait also works to the detriment of all stakeholders:
\Ni
news publishers succumbing to economic pressure undermine their established reputations and journalistic codes of ethics;
\Nii
social media platform operators increase user engagement on their networks at the expense of user experience;
\Niii
users of social media platforms unconsciously have their curiosity ``tickled'', often not realizing they are being manipulated. When they finally do realize, clickbait is rather perceived as a nuisance, much like spam.

To step out of this vicious circle, an automatic detection of clickbait, e.g., as part of ad-blockers within browsers, would enable users to opt out. The more users choose to opt out, the less effective clickbait becomes, thus forcing publishers and social networks to find other means of reaching their respective audience. Deployed at the social network itself, clickbait detection technology may unfold its impact much more rapidly. A necessary prerequisite, however, is machine learning technology capable to reliably detect clickbait.

To raise awareness and to build a community of researchers interested in this new task, we decided to organize the Clickbait Challenge~2017. In what follows, after a review of related work in Section~\ref{related-work}, we outline the challenge design and the evaluation results obtained from its first edition. Regarding its design, in Section~\ref{task} we argue for our decision to cast clickbait detection as a regression task to measure clickbait strength; in Section~\ref{dataset}, the Webis Clickbait Corpus~2017~\cite{stein:2018l} is reviewed, which was used as evaluation dataset for the challenge; and in Section~\ref{organization}, organizational details of the challenge are discussed. Section~\ref{results} presents the submitted approaches and reports on their achieved performance. We conclude with a description of how, even after the~2017 challenge has passed, researchers can still use our evaluation-as-a-service platform TIRA to evaluate clickbait detection algorithms against the ones submitted before.

\section{Related Work}
\label{related-work}

\begin{table*}[tb]%
\centering%
\fontsize{9pt}{10.5pt}\selectfont%
\setlength{\tabcolsep}{8.25pt}%
\renewcommand{\arraystretch}{1.5}%
\caption{Overview of existing clickbait corpora. All are in English.}%
\label{table-clickbait-corpora}%
\vspace{-1ex}%
\begin{tabular}{@{}lcccr@{}}
\toprule
\bfseries Publication & \bfseries Annotation Scale & \bfseries Teaser Type & \bfseries Article Archival & \bfseries Size \\
\midrule
\citet{agrawal:2016}       & Binary & Headline & No  &  2,388 \\
\citet{stein:2016b}        & Binary & Tweet    & Yes &  2,992 \\
\citet{biyani:2016}        & Binary & Headline & No  &  4,073 \\ % Not public.
\citet{chakraborty:2016}   & Binary & Headline & No  & 15,000 \\
\citet{rony:2017}          & Binary & Headline & No  & 32,000 \\
\midrule
\bfseries \citet{stein:2018l} & Graded & Tweet & Yes & 38,517 \\
\bottomrule
\end{tabular}%
\end{table*}

To the best of our knowledge, Table~\ref{table-clickbait-corpora} lists all clickbait-related datasets that have been published to date. In the last row, our Webis Clickbait Corpus~2017~\citep{stein:2018l} is listed, which has been compiled and used for the Clickbait Challenge~2017. In the columns of Table~\ref{table-clickbait-corpora}, apart from the respective publication, the datasets are classified with respect to the annotation scale at which clickbait is assessed, the type of teaser message used in the dataset, whether the articles referred to in the teasers are included in the dataset, as well as the datasets' size. Comparing our clickbait challenge dataset with the others along these attributes, it becomes apparent that our dataset is the first one that measures clickbait on a graded scale, a decision further motivated in Section~\ref{task}. Other than this, the clickbait challenge dataset adopts the construction principles of our previous Webis Clickbait Corpus~2016~\citep{stein:2016b}, while providing for an order of magnitude more training examples. Its construction is summarized in Section~\ref{dataset}.

\enlargethispage{-3\baselineskip}
All except the last publication listed in Table~\ref{table-clickbait-corpora} also propose an approach for the automated detection of clickbait alongside their dataset. An additional clickbait detection approach is presented by \citet{anand:2017}, who use the dataset of \citet{chakraborty:2016} for evaluation. In what follows, we highlight the differences of the published approaches concerning the features used for clickbait detection, as well as the (best-performing) classification technology applied. The same analysis is repeated in Section~\ref{results} for the approaches submitted to the clickbait challenge.

Since rather different datasets have been used in the literature for evaluation, the respective approaches can hardly be compared performance-wise---a strong incentive for the organization of the clickbait challenge. In terms of the F1-measure, clickbait detection performance scores around~0.75 (\cite{stein:2016b,agrawal:2016,biyani:2016}) and~0.95 (\cite{chakraborty:2016,rony:2017,anand:2017}) have been reported. Given the deficiencies of the used datasets~\cite{stein:2018l}, these scores must be taken with a grain of salt, though.

Features for clickbait detection can be derived from three sources: the teaser message, the linked article, and metadata for both. While all reviewed approaches derive features from the teaser message, the linked article and the metadata are considered only by \citet{stein:2016b} and \citet{biyani:2016}, who use variants of decision trees (random forest and gradient boosted decision tree, respectively) as their classifiers. Examples of features derived from the linked article are text readability scores and the semantic similarity of the teaser with (parts of) the article. Examples of metadata-based features are the publisher name, whether the teaser contains multimedia content, and the frequency of specific characters in the article~URL. As for features derived from the teaser message itself, the above two approaches, as well as that of \citet{chakraborty:2016} (their best performance comes from an SVM~classifier), use a multitude of structural, word-level, n-gram, and linguistically motivated features. The remaining approaches are based on embeddings as feature representation of the teaser message. \citet{agrawal:2016} takes pre-trained word embeddings as input for a~CNN, which are further updated during the training of the classifier. \citet{anand:2017} use both character and word embeddings to train an~RNN. Lastly, \citet{rony:2017} average the sub-word embeddings of a teaser message to obtain a feature representation which is then used to train a linear classifier (not further specified).

\section{Clickbait Strength Regression}
\label{task}

Previous work considered clickbait as a binary phenomenon: either a teaser message is clickbait or it is not. Though there are messages that are obviously clickbait (``You won't believe what happened!''), we noticed that making a decision is often not as straightforward. The reason for this is that, since the very purpose of teaser messages is to attract the attention of readers, every message containing a link baits user clicks to some extent. The question is whether this baiting is perceived as immoderate or deceptive by the reader---a subjective notion. To work around this subjectivity we asked how heavily a teaser message makes use of clickbaiting techniques, leaving open the question whether it is clickbait.

\begin{figure*}%
\includegraphics[width=0.245\textwidth]{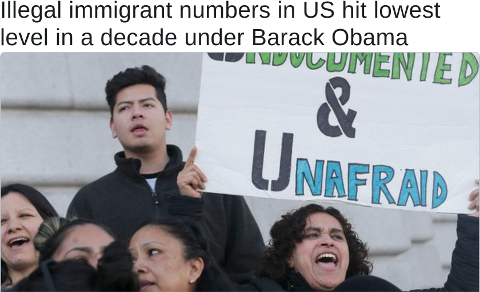}\hfill%
\includegraphics[width=0.245\textwidth]{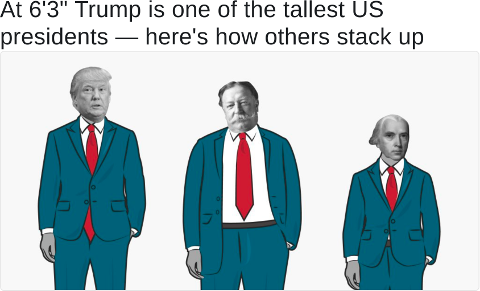}\hfill%
\includegraphics[width=0.245\textwidth]{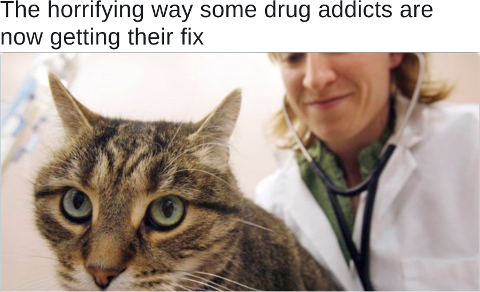}\hfill%
\includegraphics[width=0.245\textwidth]{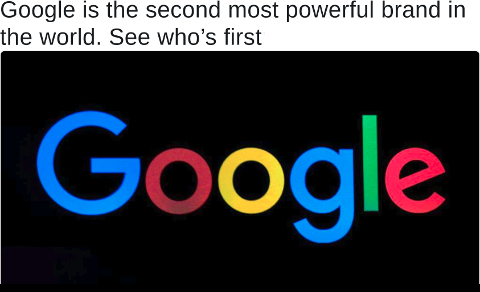}%
\vspace{-1ex}%
\caption{Examples of teaser messages from news publishers on Twitter (the more to the right, the more clickbaiting). The tweets have been redacted, removing information identifying the publishers in order not to pillory them.}%
\label{clickbait-examples}%
\end{figure*}

The teaser messages depicted in Figure~\ref{clickbait-examples} serve as an illustration of the varying degrees of clickbait strength. The messages are arranged by increasing clickbait strength  from left to right as judged by many assessors. The first teaser message on the left makes a clear statement about the information conveyed in the linked article (links have been omitted), and likely only very few readers would consider this message clickbait. In the second teaser message, the text of the message is comparably more clickbaiting, but the image below provides valuable additional clues (the image ``spoils'' the information absent from the text message). The same is true for the third teaser message; however, the image only conveys a very vague idea about the information missing from the teaser text. Whether these two tweets should be classified as being clickbait is difficult to decide, but they are obviously more clickbaiting than the first one. The fourth teaser message on the right finally makes heavy use of clickbaiting techniques, and the vast majority of readers would classify this teaser message as a paragon of clickbait.

What this sequence of teaser messages exemplifies, and what our large-scale crowdsourcing study corroborates, is that there is a whole continuum of teaser messages between the extremes of clickbait and non-clickbait. Although it is apparent that teaser messages that fall in-between could have been formulated in a more informative way to render them less clickbaiting, it is questionable whether they are clickbaiting enough to be perceived as clickbait in general. Because of this, to account for different degrees of clickbait strength, we opted for casting clickbait detection as a regression problem in the clickbait challenge, and used a graded scale to annotate the teaser messages in our dataset. The graded scale has four values (Likert scale with forced choice) and ranges from ``not'' via ``slightly'' and ``considerably'' to ``heavily'' clickbaiting.

\section{Evaluation Dataset}
\label{dataset}

To provide a high-quality, representative evaluation dataset for the clickbait challenge, we compiled the Webis Clickbait Corpus~2017. The corpus is authentic, representative, rich in terms of potential features, unbiased, and large-scale. Since the dataset has been described at length by \citet{stein:2018l}, we only summarize the most important points here.

Table~\ref{table-clickbait-corpus} gives an overview of the main characteristics of our acquisition and annotation process. We build on and extend the approach taken to construct the smaller Webis Clickbait Corpus~2016 \cite{stein:2016b}, rendering both corpora comparable.

\begin{table*}[tb]%
\caption{Webis Clickbait Corpus 2017: Corpus acquisition overview (left), corpus annotation overview (right).}%
\label{table-clickbait-corpus}%
\fontsize{9pt}{10.5pt}\selectfont%
\setlength{\tabcolsep}{12pt}%
\renewcommand{\arraystretch}{1.25}%
\vspace{-1ex}%
\begin{tabular}[t]{@{}ll@{}}
\toprule
\multicolumn{2}{@{}c@{}}{\bfseries Corpus Acquisition} \\
\midrule
Platform:          & Twitter \\
Crawling period:   & Dec 1 2016 -- Apr 30 2017 \\
Crawled tweets:    & 459,541 \\
\addlinespace
Publishers:        & \parbox[t]{0.25\textwidth}{\raggedright 27 \quad \color{darkgray} (abc, bbcworld, billboard, bleacherreport, breitbartnews, business, businessinsider, buzzfeed, cbsnews, cnn, complex, espn, forbes, foxnews, guardian, huffpost, independent, indiatimes, mailonline, mashable, nbcnews, nytimes, telegraph, usatoday, washingtonpost, wsj, yahoo)} \\
\addlinespace
Filters:           & - No videos in tweets. \\
                   & - Exactly one hyperlink in tweet. \\
                   & - Article archiving succeeded. \\
                   & - Main content extraction succeeded. \\
\addlinespace
Recorded fields:   & \parbox[t]{0.25\textwidth}{\raggedright 12 \quad \color{darkgray} (postId, postTimestamp, postText, postMedia, postPublisher, targetUrl, targetTitle, targetDescription, targetKeywords, targetParagraphs, targetCaptions, targetWarc\-Archive)} \\
\addlinespace
Sampling strategy: & Maximally 10 tweets per day and publisher \\
Sampled tweets:    & 38,517 \\
%\addlinespace
\bottomrule
\end{tabular}%
\hfill%
\begin{tabular}[t]{@{}ll@{}}
\toprule
\multicolumn{2}{@{}c@{}}{\bfseries Corpus Annotation} \\
\midrule
Crowdsourcing Platform: & Amazon Mechanical Turk\\
Annotations per tweet: & 5\\%\parbox[t]{0.2\textwidth} 5 \\
Annotation Scheme: &  \parbox[t]{0.25\textwidth}{\raggedright 4-point Likert scale. \\ \color{darkgray} Values: Not clickbaiting (0.0), Slightly clickbaiting (0.33), Considerably clickbaiting (0.66), Heavily clickbaiting (1.0) }\\
\addlinespace
\multicolumn{2}{@{}l@{}}{Mode distribution of the annotations incl.\ agreement levels:}\\
\multicolumn{2}{@{}c@{}}{\includegraphics[width=0.4\textwidth]{mode_distributions2}}\\
\bottomrule
\end{tabular}%
\end{table*}

As teaser type for our corpus, we chose Twitter tweets since
\Ni
the platform has a large user base, and
\Nii
virtually all major US~news publishers disseminate their articles through Twitter.
Our sample of news publishers is governed by publisher importance in terms of retweets. Restricting ourselves to English-language publishers, we obtain a ranking of the top-most retweeted news publishers from the NewsWhip social media analytics service.%
\footnote{https://www.newswhip.com}

Taking the top~27 publishers, we used Twitter's API to record every tweet they published in the period from December~1,~2016, through April~30,~2017. To enable the research community to develop and experiment with a rich set of features, we included the tweet text, media attachments, and the metadata provided by Twitter.Furthermore, we crawled the news article advertised using the Webis Web Archiver~\cite{stein:2018v}, which records the whole communication that takes place between a client (browser) requesting the web page of a news article and the publisher's web server hosting it, storing it in web archive (WARC) files (including, e.g., HTML, CSS, Javascript, and images). This way, every article page that forms part of our corpus can be reviewed as it was on the day we crawled it, allowing for corpus reviews even after years, hence maximizing its reproducibility. Nevertheless, users of our corpus will not have to handle the raw WARC files. For convenience, we applied publisher-specific wrappers extracting a set of content-related fields (cf.\ fields prefixed with ``target'' in Table~\ref{table-clickbait-corpus}).

To obtain a sample of tweets that has a high topic diversity, we crawled news as described above for five months in a row, yielding almost half a million tweets that fit our criteria and that were successfully archived. From this population, we drew a random sample for annotation where, for budgetary reasons, the goal was to draw at least 30,000 tweets and at most 40,000. Since the distribution of tweets per publisher is highly skewed, we apply stratified sampling to avoid a corresponding publisher bias. Similarly, we ensure that tweets are sampled from each day of the five months worth of tweets to cover the whole time period. Selecting a maximum of ten tweets per day and publisher yielded a set of 38,517 tweets and archived articles, which were then subjected to manual annotation.

The annotation of the tweets regarding clickbait strength was implemented with the crowdsourcing platform Amazon Mechanical Turk (AMT). For each tweet, we requested annotations from five different workers. To guarantee a high-quality dataset, all crowdsourced assessments were reviewed and, if necessary, discarded, resubmitting the respective assignment to AMT.

The histogram in Table~\ref{table-clickbait-corpus} shows the distribution of tweets across the four classes of our graded scale as stacked bars. To classify a tweet into one of the four classes, the mode of its annotations is used, where, in case of multiple modes, the fifth annotation is used as a tie breaker. The different colors in the bars encode different levels of agreement. With a value of~$0.21$ in terms of Fleiss'~$\kappa$, the annotator agreement is between slight and fair. However, when binarizing the classes by joining the first and last two classes into one, $\kappa$~becomes~$0.36$, which corresponds to the respective value of~$0.35$ reported for our previous clickbait corpus \cite{stein:2016b}. Furthermore, also the distribution of tweets across the binarized classes matches that of our previous corpus. Recalling that our previous corpus has been assessed by trained experts, we conclude that our crowdsourcing strategy lives up to the state of the art and that it can be considered as successful: the two independently designed and operationalized annotation studies still achieve the same result, hence our annotation experiment can be understood as a reproduction of our previous efforts, only at a larger scale. 

\section{Challenge Organization}
\label{organization}

To organize and manage the clickbait challenge, we use the eval\-uation-as-a-service platform TIRA~\cite{stein:2012k,stein:2014j},%
\footnote{\url{https://tira.io}}
which provides the means to host competitions that invite software submissions. In contrast to run submissions, challenge participants get access to a virtual machine on which they deploy (=submit) their clickbait detection approach. The deployed approach is evaluated on a test dataset hosted at TIRA, so that participants cannot gain direct access to it, giving rise to blind evaluation. For task organizers, this procedure has the advantage that it allows, given the consent of the authors, to reevaluate the submitted clickbait detection approaches also on new datasets (since the approaches are maintained in executable form in the virtual machines), and to evaluate future clickbait detection approaches in a meaningful way (since the test data are kept private).

To participate in the clickbait challenge, teams had to develop regression technology that rates the ``clickbaitiness'' of a social media post on a $[0,1]$ range (a value of~1 denoting heavy clickbaiting). For each post, the content of the post itself as well as the main content of the linked target web page are provided as JSON-Objects in our datasets. As primary evaluation metric, mean squared error~(MSE) with respect to the mean judgment of the annotators is used. For informational purposes, we also compute the F1~score with respect to a binarized clickbait class (from the mode of the judgments), and we measure the runtime of the classification software.

We published 19,538~tweets from our dataset as a training dataset for the challenge participants, and kept the remaining 18,979~tweets for the private test dataset. As a strong baseline for the challenge, we used a modified version of our own seminal classifier \cite{stein:2016b}. To account for the recast of clickbait detection as a regression task, we used the same feature set but replaced the random forest classifier with a ridge regression algorithm. A rather weaker baseline just predicts the average true clickbait score of the test data for every tweet.

\newpage
\section{Submitted Approaches and Results}
\label{results}

\enlargethispage{-2\baselineskip}
From the 100~teams that registered for the Clickbait Challenge~2017, 33~finally requested a virtual machine when asked. From these 33~teams, 13~teams followed through, made a successful submission, and evaluated their approach on the test dataset. Their performance results are shown in Table~\ref{table-evaluation-results}.%
\footnote{To render talking about the respective approaches more consistent (and more fun), we adopted a naming scheme for this task: each team chose a ``code name'' for their approach from a list of fish names, loosely alluding to the ``bait'' part of clickbait.}
As can be observed in the second column of Table~\ref{table-evaluation-results}, 6~of the~13 submitted approaches outperformed our strong baseline in terms of minimizing the mean squared error~(MSE), the best performance being achieved by zingel~\cite{zingel} with a mean squared error of~0.033. The runner-ups are emperor (no notebook submitted, approach uses a CNN on the teaser text) and carpetshark~\cite{carpetshark}, both achieving a mean squared error of~0.036. Furthermore, eight teams, as well as our strong baseline, outperform the weak baseline. This can be seen from the normalized mean squared error~(NMSE) scores in the third column of Table~\ref{table-evaluation-results}, where the mean squared error achieved by a clickbait detection approach is divided by the weak baseline's performance (MSE=0.0735). Hence, an NMSE~score less than~1 means that the approach outperforms the weak baseline.

\begin{table*}[tb]%
\caption{Performance results achieved by the approaches submitted to the Clickbait Challenge~2017.}
\label{table-evaluation-results}
\begin{center}
\begin{small}
\renewcommand{\tabcolsep}{10pt}
\renewcommand{\arraystretch}{1.2}
\vspace{-1ex}
\begin{tabular}{@{}lccccccrl@{}}
\toprule
Team                &MSE  &NMSE &F1   &Prec &Rec  &Acc  &Runtime  &Publication\\
\midrule
zingel              &0.033&0.452&0.683&0.719&0.650&0.856&00:03:27 &\citet{zingel}\\
emperor             &0.036&0.488&0.641&0.714&0.581&0.845&00:04:03 & --\\
carpetshark         &0.036&0.492&0.638&0.728&0.568&0.847&00:08:05 &\citet{carpetshark}\\
arowana             &0.039&0.531&0.656&0.659&0.654&0.837&00:35:24 & --\\
pineapplefish       &0.041&0.562&0.631&0.642&0.621&0.827&00:54:28 &\citet{pineapplefish}\\
whitebait           &0.043&0.583&0.565&0.699&0.474&0.826&00:04:31 &\citet{whitebait}\\
clickbait17-baseline&0.044&0.592&0.552&0.758&0.434&0.832&00:37:34 &\citet{stein:2016b} \\
\midrule
pike                &0.045&0.606&0.604&0.711&0.524&0.836&01:04:42 &\citet{pike}\\
tuna                &0.046&0.621&0.654&0.654&0.653&0.835&06:14:10 &\citet{tuna}\\
torpedo             &0.079&1.076&0.650&0.530&0.841&0.785&00:04:55 &\citet{torpedo}\\
houndshark          &0.099&1.464&0.023&0.779&0.012&0.764&00:26:38 & --\\
dory                &0.118&1.608&0.467&0.380&0.605&0.671&00:05:00 & --\\
salmon              &0.174&2.389&0.261&0.167&0.593&0.209&114:04:50&\citet{salmon}\\
snapper             &0.252&3.432&0.434&0.287&0.893&0.446&19:05:31 &\citet{snapper}\\
\bottomrule
\end{tabular}
\end{small}
\end{center}
\end{table*}

In terms of the F1~measure (fourth column), the six top approaches and three others outperform the strong baseline; a result at least partly rooted in the fact that some of the approaches have been optimized with respect to F1 instead of~MSE. In terms of~F1, the top-performing approach is again zingel. Together with the observation that this approach is also the fastest one (eighth column), this underlines its high quality. Regarding precision and recall individually (columns six and seven), one can observe that the approaches outperforming the baseline in terms of~F1 succeed by trading small losses in precision with significant gains in recall.

Of the~13 teams that made a successful submission, 9~also submitted a notebook paper, referenced in the last column of Table~\ref{table-evaluation-results}. In what follows, we briefly summarize each approach.

Zingel by \citet{zingel} is the best-performing approach of the Clickbait Challenge~2017. It employs a neural network architecture with bidirectional gated recurrent units~(biGRU) and a self-attention mechanism to assess clickbait strength (namely, the mean of the clickbait annotations for a tweet). As input to the network, only the teaser text is considered, which is represented as a sequence of word embeddings that have been pre-trained on Wikipedia using Glove (but then updated during training).

Carpetshark by \citet{carpetshark} employs an ensemble of SVM~regression models (so-called extremely randomized forests), one each trained separately for the text fields provided in the training data. Besides the teaser text, these fields are the keywords, descriptions, image captions, paragraphs, and the title of the linked article. In addition to the objective of predicting clickbait strength, predicting the standard deviation of the annotations is considered, improving the performance of the ensemble. An attempt at augmenting our challenge dataset with own data failed to improve the approach's performance and was hence omitted from the final submission.

Pineapplefish by \citet{pineapplefish} relies on a ``linguistically infused'' neural network with two sub-networks. An LSTM~sub-network that takes as input a sequence of 100~word embeddings (pre-trained on a Twitter dataset and then updated during training). The sequence consists of the first 100~content words of the teaser text, and, in case the teaser text is shorter than 100~content words, text from the linked article's fields. The second sub-network consists of two dense layers which take as input a vector of linguistic features extracted from the teaser message and the linked article text. The authors also experimented with object detection technology to obtain features from teaser images, however, these features were not included in their final model due to a lack of performance gain.

Whitebait by \citet{whitebait} analyzes all text fields available for each training example (=tweet), as well as the publication time of the tweet. For each of the text fields, an LSTM is trained that takes as input a sequence of word embeddings (initialized randomly). For the publication date, a neural network with one hidden layer is trained. As input, the publication time is binned into one-hour ranges and then converted into one-hot encodings. In a second step, the individually trained networks are fused by concatenating the last dense layer of the individual networks. The authors state that this two-step procedure performs better than training a complete model from scratch. Attempts at exploiting the teaser images have not made it into the final version of this model.

Pike by \citet{pike} computes a set of~175 linguistic features from the teaser message, two features related to the linked article, as well as three features that capture semantic relations between the teaser message and the linked article. This feature set is then fed into a random forest regression algorithm, which achieved the best performance compared to other alternatives tested in a 10-fold cross validation on the training dataset.

Tuna by \citet{tuna} consists of a deep neural network with three sub-networks. The first sub-network is a bidirectional LSTM with attention which gets as input a sequence of word embeddings (pre-trained on a Google News dataset and updated during training) representing the teaser text. The other two sub-networks are Siamese neural networks, the first of them producing a similarity score between the teaser text and the description field of the linked article, the second one producing a similarity score between the teaser image and the target description. As representations for the teaser text and the target description, doc2vec embeddings of the fields are employed. To represent the teaser image, a pre-trained object detection network was applied to the image, and the activation on a convolutional layer was taken as image representation.

Torpedo by \citet{torpedo} uses pre-trained Glove word embeddings (on Wikipedia and a further dataset) to represent the teaser message of a tweet. For this, the word embeddings for the different words in the teaser text are averaged. In addition, seven handcrafted linguistic features are added to the representation. With this feature set, a linear regression model is trained to predict the clickbait strength of tweets.

Salmon by \citet{salmon} applies gradient boosting (XGBoost) to a tweet representation that consists of three feature types:
\Ni
teaser image-related features encoding whether there is a teaser image, and, using OCR on the image, whether there is text in the image,
\Nii
linguistic features extracted from the teaser text and the linked article fields, and
\Niii
features dedicated to detect so-called abusers that are supposed to capture user behavior patterns.

\enlargethispage{-2\baselineskip}
Snapper by \citet{snapper} trains separate logistic regression classifiers on different feature sets extracted from the teaser text, the linked article title, and the teaser images (features extracted first using the Caffe library).%
\footnote{\url{http://caffe.berkeleyvision.org}}
In a second step, the predictions of the individual classifiers are taken as input to train a final logistic regression classifier.

\section{Conclusion}

The Clickbait Challenge~2017 stimulated research and development towards clickbait detection: 13~approaches have been submitted to the challenge. Many of these approaches have been released open source by their authors.%
\footnote{We collected them here: https://github.com/clickbait-challenge}
Together with the working prototype deployed within virtual machines at TIRA, this renders the proceedings of the clickbait challenge reproducible, and newcomers have an easier time following up on previous work.

Several more approaches have been proposed and submitted to TIRA after the challenge had ended. Together with zingel, these four additional approaches are the top five best-performing clickbait detectors on the leaderboard at the time of publishing the current challenge overview.%
\footnote{https://www.tira.io/task/clickbait-detection/}
The leading approach, albacore by \citet{albacore}, like zingel, employs a biGRU~network, initialized by Glove word embeddings. The runner-up anchovy is also an adaptation of zingel, whereas icarfish by \citet{icarfish} demonstrates that our baseline \cite{stein:2016b} is still competitive: when optimizing the selection of features using a newly proposed feature selection approach, the baseline approach improves substantially. For the two approaches anchovy and ray, at the time of writing, no written reports have surfaced. More teams have registered after the first challenge has passed, now working on new approaches to solve the task. We will keep the evaluation system running for as long as possible to allow for a continued and fair evaluation of these new approaches.

\begin{raggedright}

\end{raggedright}
\end{document}